\documentclass{article}
\usepackage{ijcai21}

\usepackage{times}

\usepackage{soul}
\usepackage{url}
\usepackage[hidelinks]{hyperref}
\usepackage[utf8]{inputenc}
\usepackage[small]{caption}
\usepackage{graphicx}
\usepackage{amsmath}
\usepackage{booktabs}
\usepackage{epstopdf}
\title{ConsciousControlFlow(CCF): Conscious Artificial Intelligence based on Needs}

\author{
Hongzhi Wang$^1$\footnote{Contact Author}\and
Bozhou Chen$^1$\and
Yueyang Xu$^1$\And
Kaixin Zhang$^1$\And
Shengwen Zheng$^1$\\
\affiliations
$^1$Harbin Institute of Technology, Harbin, Heilongjiang, China\\
\emails
\{wangzh, bozhouchen\}@hit.edu.cn,
1404039935@qq.com,
\{1170300216, 1170300707\}@stu.hit.edu.cn
}

\begin{document}

\maketitle

\begin{abstract}
The major criteria to distinguish conscious Artificial Intelligence (AI) and non-conscious AI is whether the conscious is from the needs. Based on this criteria, we develop ConsciousControlFlow(CCF) to show the need-based conscious AI. The system is based on the computational model with short-term memory (STM) and long-term memory (LTM) for consciousness and the hierarchy of needs. To generate AI based on real needs of the agent, we develop several LTMs for special functions such as feeling and sensor. Experiments demonstrate that the the agents in the proposed system behave according to the needs, which coincides with the prediction.
\end{abstract}

\section{Introduction}

Currently, Artificial Intelligence (AI) gains great advances. However, current focus is \emph{functional AI}, which provides some specific functions such as face recognition, the game of go and question-answer. Different from functional AI, conscious AI~\cite{DBLP:conf/aaaiss/2019tocais} aims to build AI systems with consciousness. Conscious AI will not only help to build better AI systems by solving the problem of data-driven approaches but also create the opportunity to study neuroscience and behavior science by connecting behavior with conscious activities.

With its wide applications and great interests, many researchers devote to the model of consciousness. Existing models of consciousness are in three aspects. The first is from the essentials of consciousness such as global workspace theory~\cite{baars1988,newman1993}. The second is from the attention control, a basic function of consciousness and based on attention schema theory~\cite{Graziano2014}. The third focuses on the reasoning function for the consciousness, the examples of which include Ouroboros model~\cite{Thomsen2011} and Glair architecture~\cite{Shapiro2010}.
With these models, some techniques have been developed including the one based on global workspace theory~\cite{baars2009} and the one based on attention schema theory~\cite{Boogaard2017}. However, these techniques attempts to implement the function of consciousness and fail to demonstrate real consciousness of the machine globally.




The nervous system is the key factor to distinguish animals form plants. The driver of nervous system evolution is the needs such as energy, breed and safety. Since the consciousness is from the evolution of the nervous system, it is gained from the needs. Thus, the real consciousness could be distinguished by whether it is from real need. We illustrate this point with an example. The function of the real pain is to tell body to avoid harm. This is from the need of the body and from the consciousness. Otherwise, it is not real pain.


Motivated by this, we develop a system with \underline{c}onscious \underline{c}ontrol \underline{f}low (CCF) based on needs to show the conscious AI.

\begin{figure}
	\centering
	\includegraphics[width=0.9\linewidth]{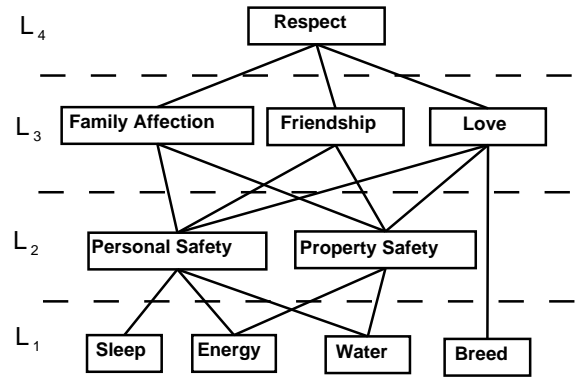}
	\caption{The Hierarchy of Needs}
	\label{fig:need}
\end{figure}

A conscious agent may have various needs in different levels. For example, the need of food is the basic need in low level, while the need of love is the need in higher level. To represent the needs effectively, we organize them in hierarchy structure, inspired by Maslow's hierarchy of needs~\footnote{\url{https://www.simplypsychology.org/maslow.html}}.  In such structure, a need in higher level is the prediction of the lower-level needs. For example, as shown in Figure~\ref{fig:need}, the needs in the lowest level are the basic needs such as energy, water and sleep. The need of personal safety is the higher need, which is the prediction of sleep, energy and water.

For an agent, multiple needs may emerge in the same time, which may be sensed and processed unconsciously, while the consciousness runs in series~\cite{baars2009,DBLP:journals/corr/abs-2011-09850}. Thus a competitive mechanism is in demand to select the most valuable stuff for the conscious to handle. To model the consciousness, the unconscious behaviors as well as the competitive mechanism, our system adopts short-term memory and long-term memory (STM-LTM) architecture according to the discovery of neuroscience~\cite{baars1988}. In the architecture, a STM represents the consciousness with limited space, and various LTMs implements complex real mantel functions such as pain, body control, natural language processing, vision. LTMs compete to enter STM according to their strengths. The strengths are determined by the needs. Once an LTM enters STM, it comes to the consciousness.

The contributions of this paper are summarized as follows.

\begin{list}{\labelitemi}{\leftmargin=1em}\itemsep 0pt \parskip 0pt
	
	\item We discover the connection between the hierarchy of needs and consciousness and apply such discovery for a conscious AI.
	
	\item We develop a system for the conscious AI. In this system, we construct the STM-LTM framework and develop some typical LTMs. The system is based on the hierarchy of needs which drive the competition of LTMs. To the best of our knowledge, this is the first work for need-driven conscious AI.
	
	\item To illustrate the effectiveness of the proposed techniques, we design typical scenarios and conduct experiments on it with conscious agents. The experimental results demonstrate that our system could act as prediction with explainable mental behaviors, which show that our system gain conscious AI in some level.
\end{list}

The remaining part of this paper is organized as follows. Section~\ref{sec:model} introduces the computational model of consciousness with the hierarchy of needs. Section~\ref{sec:tech} proposes the techniques for the system including system architecture and algorithms. Section~\ref{sec:exp} presents experimental results with analysis. Section~\ref{sec:con} draws the conclusions.



\section{Computational Model for Consciousness}
\label{sec:model}

We believe in that consciousness is the product of evolution which leads to better satisfaction of needs. Thus, we distinguish conscious AI to non-conscious AI in whether the decision is from the needs of the individuals.

We base system on the hierarchy of needs. To simplify the system, we implement four levels of needs as shown in Figure~\ref{fig:need}. As the base, each individual in our system has 4 basic needs, i.e. sleep, energy, water and breed, which are quantized as a \emph{state vector}, respectively. The need in the higher level is considered as a prediction with needs in lower level.

\begin{figure}
	\centering
	\includegraphics[width=.25\textwidth]{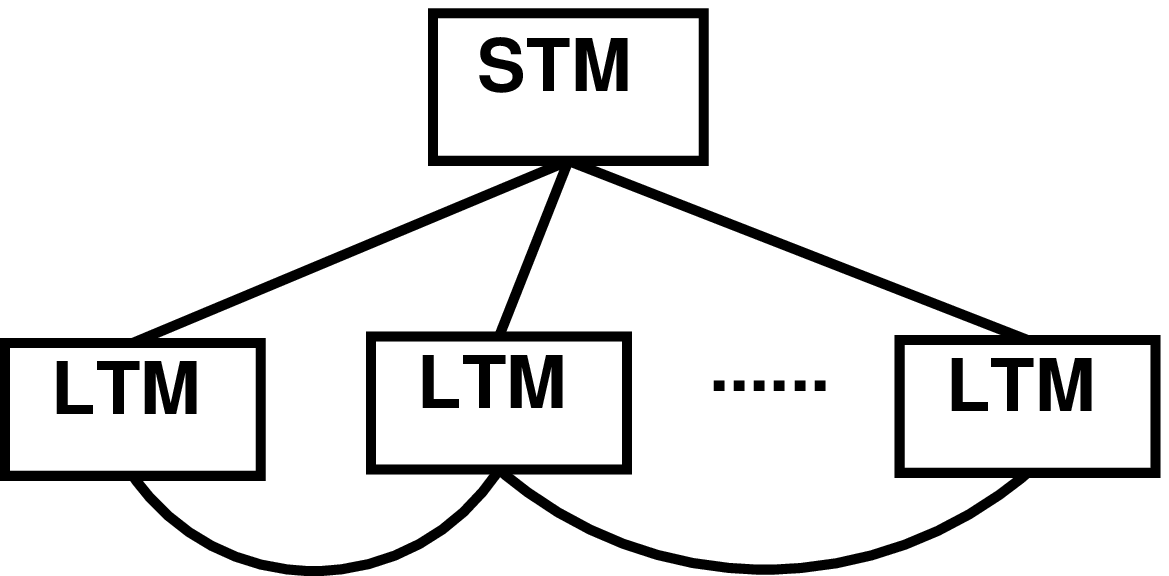}
	\caption{The Computational Model for Consciousness}
	\label{fig:model}
\end{figure}

To implement the consciousness, we adopt the computational model of consciousness~\cite{blumtalk} inspired by neuroscience. The model is illustrated in Figure~\ref{fig:model}. In this model, each long-term memory (LTM) is a processor with memory in charge of a function such as speech, face recognition or angry. LTMs have connection to each other and connect to the short term memory (STM) with limited space, i.e. 7$\pm$2 slots~\footnote{\url{https://en.wikipedia.org/wiki/The_Magical_Number_Seven,_Plus_or_Minus_Two}}. STM is the consciousness and makes conscious decisions. STM has no awareness of how the unconscious LTM works.

\section{Techniques}
\label{sec:tech}

\begin{figure}
    \centering
    \includegraphics[width=.45\textwidth]{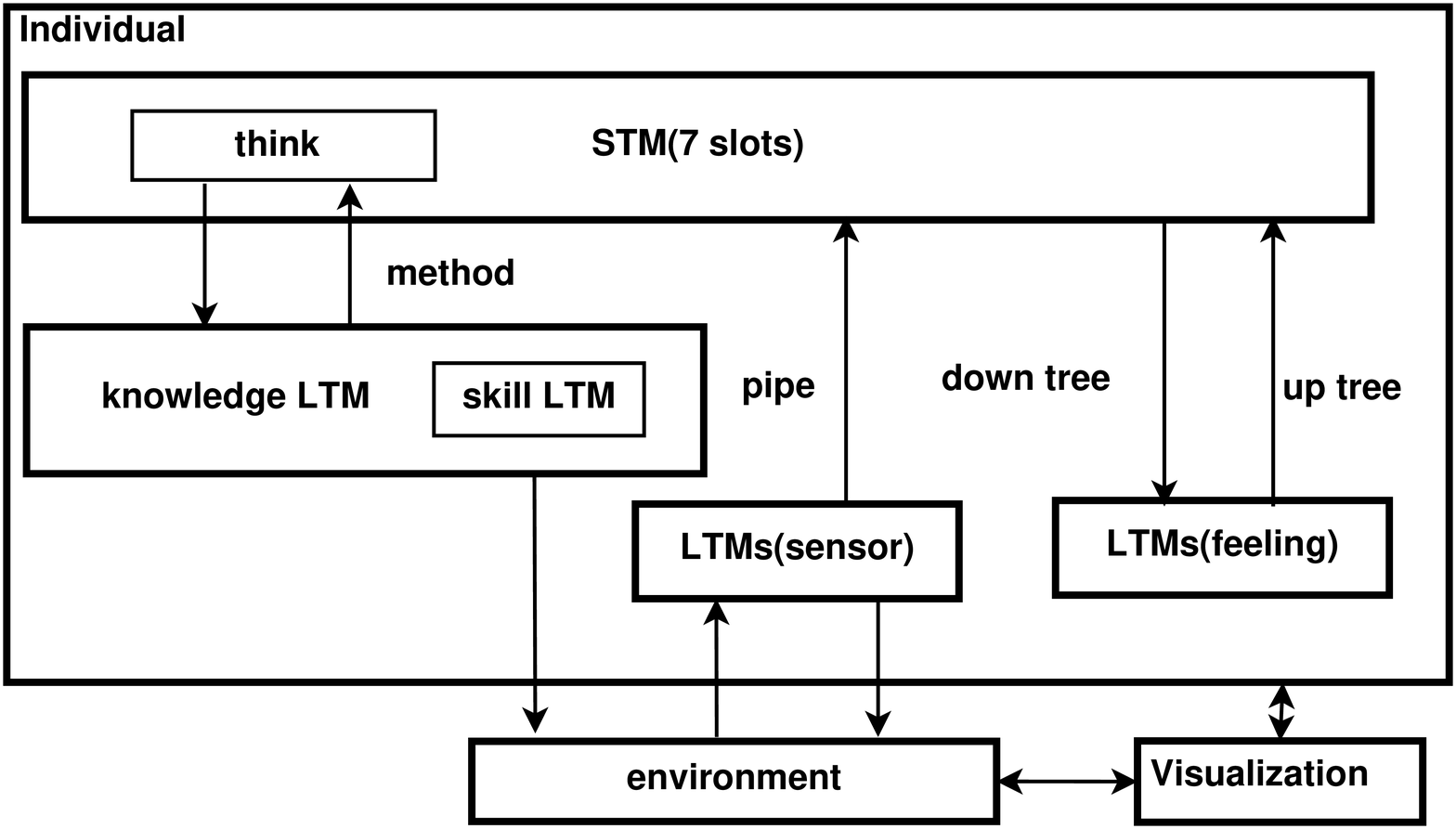}
    \caption{The Architecture of CCF}
    \label{fig:arch}
\end{figure}

According to the computational model, the components of the system are shown in Figure~\ref{fig:arch}. The components are introduced as follows.

\noindent \textbf{Individual} This is the core component of the system, in which the consciousness is implemented. An individual is compounded with a STM, multiple LTMs and links, which will be discussed in Section~\ref{sec:STM}, Section~\ref{sec:ltm} and Section~\ref{sec:link}, respectively.


\noindent \textbf{Environment} This component controls the environment such as temperature, available food and water, which may possibly affect the behavior. The change of environment could be performed randomly or input by the user.

\noindent \textbf{Visualization} This component visualizes the environment and the behaviors of individuals to generate the animation.

\subsection{STM}
\label{sec:STM}
\noindent \textbf{STM} could be considered as a central processor with a cache, whose basic unit is called slot. This is a special cache. On the one hand, those slots are organized as a tree rather than a liner structure. On the other hand, the number of slot in cache has an upper bound of 7~\cite{miller1956magical}. Further, some operations are defined for STM to take operations on the cache, and these operations are called \textsf{think} uniformly. As unit of STM's storage, slot has four types, i.e. \emph{ontology}, \emph{need}, \emph{object} and \emph{method}.

\noindent \textbf{Slots$'$} four specific types are summarized. Except \textsf{ontology}, each type may have multiple instances. For the convenience of discussions, we give a label to each instance when it is loaded in STM to distinguish different slots with same type.

These four slot types are introduced as follows.
\begin{list}{\labelitemi}{\leftmargin=1em}\itemsep 0pt \parskip 0pt

\item \textsf{Ontology} identifies the agent itself. It has only one instance denoted as $I_S$. If the agent is conscious, $I_S$ is in STM as the root. Otherwise, if the agent is sleeping or unconscious for some reason, the cache becomes empty, with $I_S$ switched out.

\item \textsf{Need} represents the needs from feeling LTMs which will be introduced in the next section. Each instance of \emph{need} $I_T^F$ corresponds to a feeling LTM $F$. Besides the copy of $F$, $I_T^F$ has a weight to show its intensity, which is transferred from $F$ and is a key factor to decide the need to be processed by STM. The weight computation and the details of LTMs' competition for STM will be described in Section~\ref{sec:ltm}.

\item \textsf{Object} accepts information from the environment. All objects correspond to the sensor LTM. Each one represents one kind of signals from the sensor LTM.

\item \textsf{Method} denotes the methods used to solve needs in slots. However, slots just maintain a label of the method to save more storage, and the modules of the methods are in the LTM.

\end{list}

Each item in the slot has intensity value so that when the slots are full, it can be decided whether or not to replace an item and which item is to be replaced.

\begin{figure}
    \centering
    \includegraphics[width=\linewidth]{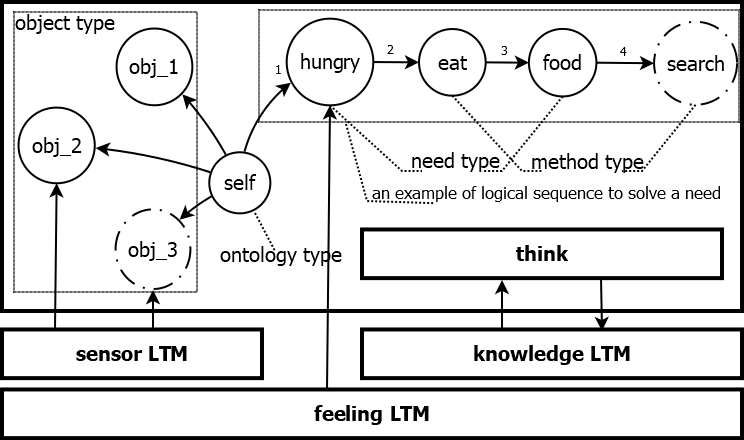}
    \caption{The internal Structure of STM}
    \label{fig:process}
\end{figure}

We use an example of solving hungry need to illustrate the work of the STM,  as is shown in Figure~\ref{fig:process}.
\begin{list}{\labelitemi}{\leftmargin=1em}\itemsep 0pt \parskip 0pt
    \item \emph{\textbf{Step 1}}. The agent $Alice$ feels hungry (accepts the hungry need from the corresponding feeling LTM). In this step, the hungry need is packed into a slot and connected to self.
    \item \emph{\textbf{Step 2}}. She tries to solve the need. The think module takes the duty to invoke knowledge LTM to find the method that is able to solve the hungry need. In this step, the method is \emph{eat} which is then packed into a slot and connected to hungry. After she thinks out a method to solve the need, she will try to conduct the method. While the truth is that she has nothing to eat (When programming, this will be a method of feasibility check).
    \item \emph{\textbf{Step 3}}. She finds that the reason that she could not conduct eating is that she does not own food. Therefore there is a new need labelled with food entering STM. It should be noted that this kind of need is different from that from feeling LTM. It can be regarded as a target here.
    \item \emph{\textbf{Step 4}}. She processes the new need for hungry, and then gets a \emph{search} method. When \emph{search} in going to enter STM, one slot must be removed from STM since the upper bound of slot's number reaches 7. In the end, \emph{obj\_3} is removed out. Then she begins searching food. In the process of searching, the object in \emph{obj\_1} and \emph{obj\_2} will accept different and new information from sensor LTM continually.
    \item \emph{\textbf{Step 5}}. She has found food. Then search is removed out. Eat method is called. After eating up, eat and food are removed out. Then everything returns to normal.
\end{list}

Note that Figure~\ref{fig:process} shows the details of STM in Figure~\ref{fig:arch}. The changes of objects in the environment are sent to the STM through sensor LTMs, which may be affected by the requirement of the agent. The STM should have the ability of handling the information from the environment and make decisions according to the changes in the environment. To achieve this goal, we abstract the information in the environment to objects, each of which is wrapped up as a package matching the slot in STM. In STM, \textsf{think} is designed to handle the such packages entered STM.




\noindent \textbf{Think} module contains many functions about thinking. It makes decisions based on knowledge LTM as that will be discussed in the next subsection. \textsf{Decision} determines the method used to solve the need. Before the method is executed, \textsf{think} detects whether the premises of the action $R$ can be met. If the conditions are satisfied, this module will calculate parameters for the action, and monitor the running of $R$ to control its ending. Otherwise, new requirements are generated by \textsf{thinking} and enter the slots.

During monitoring, \textsf{reduction}, which is one of the functions in \textbf{think} to identify the event that the need is solved, will be conducted if there is an object happening in one slot sharing the same to the label of some need. For example, if $obj_2$ is food, then reduction happens, and as a result, \emph{food} and \emph{search} will be removed from slots.

\subsection{LTMs}
\label{sec:ltm}
\noindent \textbf{Knowledge} is a special kind of LTM. The storage is in form of a knowledge graph and could be handled with graph engines.
The functions for \textsf{Knowledge} include \textsf{save}, \textsf{update} and \textsf{query}. 
\textsf{update} saves information that has been actively or passively paid attention to or repeated.
For example, just as the example before, for an agent, it actively pays attention when looking for food for tackling hunger. Therefore, such thing is stored in LTM automatically.
\textsf{query} answer the queries according to the knowledge base and return the results, e.g. make decisions according to the need that has been described above.
Apart from the basic \textsf{query} function, the agent should also has the ability to take deep thought based on knowledge and the cache to solve complex tasks, which will be discussed in future work.

\noindent \textbf{Skill} is also a special kind of LTM. We furnished skills for each need, and those skills help agent restore and interact with environment. In the future, the skills can be split and reorganized freely.  The eat and search in the above example are example skills. Basically, each need corresponds to a skill to solve it. Additionally, we develop some auxiliary skills such as search, observe, move and put some thing in some place.


Both the above two special LTMs do not directly receive internal or external information. These information is processed by other LTMs called feeling LTM and sensor LTM.

\noindent\textbf{Feeling LTMs} keep on detecting the status of individual itself. The needs are generated by them. At present, we have implemented 10 feeling LTMs including thirsty, hungry, breed, sleep, personal safety, property safety, family affection, friendship, love and respect
corresponding to the hierarchy of needs. Each feeling LTM has two basic attributes, i.e. \emph{satisfaction} and \emph{weight}. Weight evaluates the strength of the need,
and the need with the largest weight will be handled by he agent.
Satisfaction represents the degree that the need is satisfied.
We measure the satisfaction of need with the following three rules, (1). The satisfaction of each physiological need will decrease with time. (2). All the satisfactions of needs will be affected by some specific events. (3). The level of satisfaction of high-level needs has an impact on the low-level ones~\cite{taormina2013maslow}. The computation approach of needs will be introduced in Section~\ref{sec:weight}.

\noindent\textbf{Sensor LTM} keeps on detecting the status of the environment. We has two kinds of sensor LTM, one for visual information and another one for auditory information. Sensor LTM gets message from the environment directly, partitions and processes them to STM readable messages. Then sensor LTM sends these messages to STM waiting to be wrapped as a slot and processed. The transmission mechanism is implemented based on \emph{pipe}, which will be described in the next subsection.

\subsection{Links}
\label{sec:link}
The function of links is to transfer information between STM and LTMs, and also within LTMs. For the first kind of links, links could be classified into \emph{Up-Tree} and \emph{Down-Tree} according to the direction of information transformation, which are introduced as follows.

\noindent \textbf{Up-Tree} The purpose of the Up-Tree is to run competitions that determine which chunks\footnote{trunk is a copy of some LTM at a specific time} of LTM is to be loaded into STM. This part of the work refers to the CTM~\cite{basicmodel} model. The Up-Tree has a single root in STM and $n$ leaves, one leaf in each of the $n$ LTM processors.  Each directed path from a leaf to the root has the same length $h$, $h$ = $O(log n)$. Each node of the Up-Tree that is not a leaf (sits at some level $s$, $0<s \leq h$) has two children.

\noindent \textbf{Down-Tree} At each time $t$, the content of STM is broadcast via this Down-Tree to all $n$ LTMs.

Note that both Up- and Down-Trees are used to transfer internal signals. Besides Up- and Down-Trees, STM accepts the signals from the environment, which is handled with an individual component, \emph{pipe}. As shown in Figure~\ref{fig:process}, \textsf{pipe} is in charge to transfer information from sensor LTM to STM.

Note that the links described above are between STM and LTMs. The links among LTMs have been analyzed in Section~\ref{sec:ltm}.

\subsection{The Computation of Need Weight}
\label{sec:weight}

The computation of the weight of physiological needs is straightforward, since it is only negatively related to the satisfaction. The weight is computed by the maximal satisfactory minus the current one.
However, the calculation of other levels$'$ needs$'$ corresponding weight is not so easy to be expressed as a unified formula. Since we must ensure that when the requirements of the lower layers are not met, the strength of the high-level requirements cannot exceed that of the lower layers, otherwise it is wrong. At the same time, the prediction of the underlying requirements by the high-level requirements must be considered. 


When analyzing the calculation method of the weight of an LTM, we should consider two factors, one is the satisfaction corresponding to the LTM, and the other is the satisfaction corresponding to all the LTMs at the next lower level of the demand level where the LTM is located. We focus on those LTMs that are predicted by themselves. Therefore, carefully speaking, we should consider three factors as follows.

We then analyze the influence of three factors on the final weight. First, the weight represents the strength of demand. If the satisfaction is higher, then weight should be lower, so the first factor's contribution to the final weight should be negative. Secondly, the weight can be relatively large when all low-level requirements are resolved. Quantitatively, these satisfactions are positively related to weight. Finally, for those LTMs predicted by themselves, if they are not satisfied, it leads to a relatively large weight, which better satisfies these LTMs in the future. From this point of view, the third factor should be negatively related to weight. according to the above analysis, the formula for calculating weight is as follows.

\begin{eqnarray}
\label{equ:1}
\scriptsize
    w_{l,i} & = & \alpha_{l,i} * s_{l,i} + \beta_{l,i} * \sum_{j \in sons}{s_{l - 1,j}} + \nonumber \\
    && +\gamma_{l,i} * \sum_{j \in ltms\_in\_next\_layer}{s_{l - 1,j}}
\end{eqnarray}
where $w$ denotes weight, $l$ denotes the layer number, and physiology level's l is 1. $i$ and $j$ indicate some LTM in layer $l$. $s$ denotes the satisfaction of some LTM indicated by l and $j$(or $i$).

The first influencing factor is the level of satisfaction which has a strong negative correlation with the weight of needs. We call $\alpha$ the self-decreasing coefficient. 

The second factor shows the level of satisfaction of next level's needs which are predicted by this need. That is, if some lower needs predicted by this need is not satisfied well, the lower needs will be satisfied by strengthening this need. Thus, we call $\beta$ the gain factor, which is negative and reflects the influence of the third factor. For example, if one's food is provided by the parents, the need of food will enhance the need of family affection.


The third factor is the level of satisfaction of all the lower needs with positive correlation to the needs, which shows that if some lower need is not satisfied, it should be satisfied first. Thus, we call $\gamma$ the suppression coefficient, which is a positive number, corresponding to the second factor.

$\alpha,\beta,\gamma$ can be different for each LTM. In this model, we can meet the need by adjusting the value of each LTM's $\alpha,\beta,\gamma$.




In our system we can have found suitable values from Bayesian optimization with the goal as follows.

\begin{equation}
\label{equ:score}
    max(\sum_{i \in test\_sample}X_i)
\end{equation}

where $X_i = 1$ when $\arg \max_{i}\{$the weight of each LTM in sample $i\}$ = $i_{label}$.


We manually generate 100 sample data sets, and give the requirements of which LTM should be selected in these 100 cases. When all the 100 cases corresponding to a certain set of constants are successfully predicted, the values of these constants are finally determined.

While the calculation result may be a negative value, we add a positive number to the result as a correction as follows.

\begin{eqnarray}
    w_{l,i} &=& \alpha_{l,i} * s_{l,i} + \beta_{l,i} * \sum_{j \in sons}{s_{l - 1,j}} + \nonumber \\
    &&+\gamma_{l,i} * \sum_{j \in ltms\_in\_next\_layer}{s_{l - 1,j}} + \Delta
\end{eqnarray}

\section{Experiments}
\label{sec:exp}

To verify the effectiveness of the proposed model and framework, we develop the system of agents with conscious AI and conduct experimental study.

\subsection{Experimental Setting}
To verify the proposed approaches, we develop a proper scenario, which is simple enough for the explanation of the connection with mental and physical behaviors, and scalable enough to demonstrate the decisions under various environments.

\noindent \underline{Scenario Description} The basic setting is to build two conscious agents named Alice and Bob, who are friends. Only one predator threatens the agents, and agents have one food, prey, which could be obtained with some skill called hunting. Even though the scenario looks simple, 6 needs in 3 levels are related to it, i.e. sleep, energy, water, bread, safety and friendship. The methodology of the experiments is to test whether the behaviors of agents coincide to the prediction of human. Since the prediction is natural and straightforward, we make behavior prediction by the authors.


\noindent \underline{Parameters} The parameters in Formula (\ref{equ:1}), i.e. $\alpha$, $\beta$, $\gamma$, are set as follows. Since for a creature, the ``parameters'' are optimized by evolution, we also optimize the parameters according the survive possibility. We run an agent with various parameters to observe her survive time denoted by $s$. Thus, we obtain a set $S$=\{$\alpha_i$, $\beta_i$, $\gamma_i$, $s_i$\}. With sophisticated deep network, a function $\frac{1}{s}=f(\alpha, \beta, \gamma)$ is constructed to fit $S$. Then, with hyperparameter optimization techniques~\cite{DBLP:journals/corr/abs-2003-02446}, the optimal parameters are obtained.

\noindent \underline{Needs} We measure the strength of need $n_i$ as $w_i=10-sat_i$, where $\delta$=10, and $sat_i$ is the satisfaction of $n_i$. The computation of $sat_i$ is shown as follows. The initial values of all satisfaction in the first level, i.e. sleep, energy, water, breed, is set to 5. The satisfaction decrease with time steadily. We set the threshold as 5. When the satisfaction gets smaller than 5, the corresponding need arises. When the satisfaction of some need in the first level gets to 0, the agent dies. The satisfaction of the safety in the second level is measured by the distance between the agent and the predictor. The closer between the agent and the predicator, the less the satisfaction of the agent is. The threshold of safety satisfaction is also 5. The satisfactory of friendship in the third level is measured by the distance between the friend and the predator. A small distance means that the friend has more possibility in danger and a smaller friendship satisfactory. Note that the principle of the constant setting is sufficient to simulate the need of agents and the decision to the needs. Even though many parameter combinations satisfy the requirement, 10 and 5 are sufficient to achieve the goal of our experiment.


We conduct two experiments, the first one with single agent and the second one with double agents. In the remaining of this section, we show the results and analysis.

\subsection{Experiment with Single Agent}
\begin{figure}
\begin{minipage}[t]{0.49\linewidth}
    \centering
    \includegraphics[height=0.1\textheight,width=\textwidth]{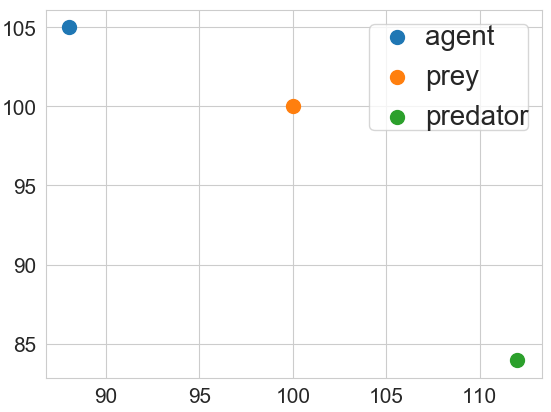}
    \caption{The Experiment with single Agent}
    \label{fig:init1}
\end{minipage}
\begin{minipage}[t]{0.49\linewidth}
    \centering
    \includegraphics[width=\textwidth]{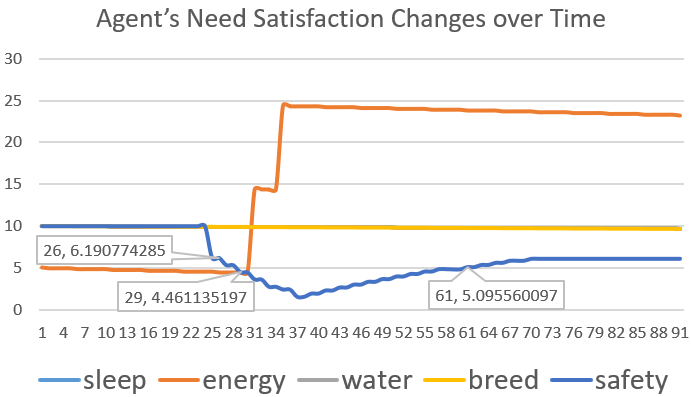}
    \caption{The Satisfaction in single-agent Experiment}
    \label{fig:exp1}
\end{minipage}
\end{figure}

This experiments have three objects, the agent Alice, prey $F$ and predators $P_1$ and $P_2$. The agent Alice is conscious. $F$ is the food of the agent and static. The predator $P$ is non-conscious. When an agent $A$ enters its view, $P$ move to $A$ for predating. The whole view is a 2-dimension space. The coordinates of these two predators are both (112, 84), and the view of predicates is the circle with them as the center and PREDATOR\_R = 23 as the radius. The speed of $P_1$ and $P_2$ is 1. The initial positions of the prey and Alice are (100,100) and (88, 105), respectively. The radius of agent view is $23$. When the prey and predator enter the view of the agent, she will acts. To accelerate the experiment, the satisfaction is initialized as 5.1. The initial scenario is shown in Figure~\ref{fig:init1}.

According to the predication based on human, Alice will move to the prey when she has the need of food. During the moving to the prey, she will also noting the predicator. The need of food drives Alice move to the prey. After preying, the safety need comes to STM, dominates the behavior of Alice and drives agent keep away from the predator.

\begin{figure}[t]
    \centering
    \includegraphics[width=0.5\textwidth]{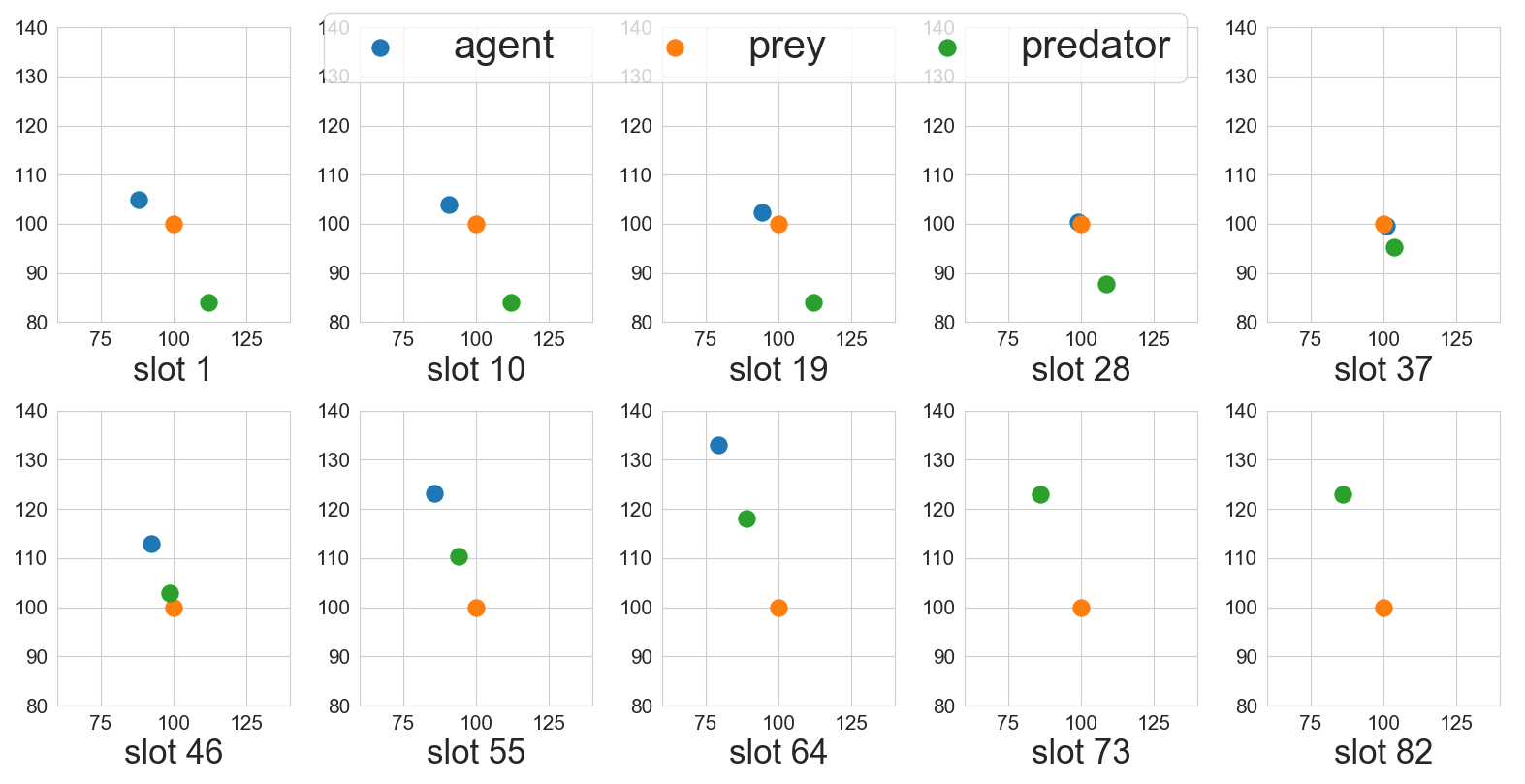}
    \caption{The Status of some Slots in single-agent Experiment}
    \label{fig:exp1full}
\end{figure}

We run the agents with 90 time slots and draw the status of some key slots in Figure~\ref{fig:exp1full}. From the results, three accidents happen. Alice starts to move to the prey based on the food need. Although Alice notices the predator, she still moves to the prey. After the preying, she starts to keep away from the predator. The behavior of the whole process coincides to the prediction.

We print the satisfactions of various needs in Figure~\ref{fig:exp1}. Note that the strength of need is evaluated with the satisfaction, the satisfaction and need weight are negative correlated.

From the figure, we observe that the food need gets below the threshold and keep decreasing. In the 32nd slot, Alice finishes preying and obtains energy. Before that, in the 28th slot, the safety need reaches the threshold, but the food need is stronger. As a result, hungry occupies the conscious and Alice attempts to finish preying. When preying is accomplished, safety becomes the most important need, and Alice starts to keep away from the predator. In the 62nd slot, Alice gets away from the predator. Her safety satisfaction gets back to the normal range and increases slowly.

From the figure, in the 32th time slot, energy increases again. According to the common sense, the safety need should be handled when the safety need reaches the lowest point. However, Alice still chooses to handle the food need. This demonstrates the feature of our framework. This is related to the uptree of the agent, which is a binary tree. The backup of the food need are stored in some medial node. During the uptree updating, the backup keeps goting up until the root is reached. Then, they come to STM and are consumed.

\subsection{Experiment with Double Agents}

In this experiment, Bob, another agent and the friend of Alice, sees the scenario in the experiment and has some mental change. The coordinate of Bob is (88, 110). To show the impact of Bob, the view radius of Alice is reduced to 16, and that of Bob is set to 35. The view of Bob is larger and could find the predator early.

\begin{figure}[t]
    \centering
    \includegraphics[width=0.49\textwidth]{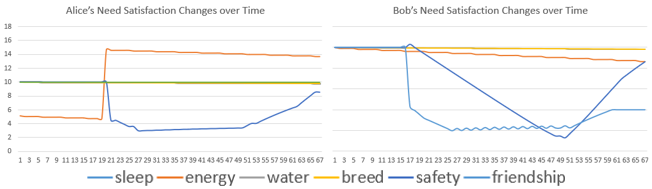}
    \caption{The Satisfaction in double-agent Experiment}
    \label{fig:exp2}
\end{figure}

\begin{figure}[t]
    \centering
    \includegraphics[width=0.5\textwidth]{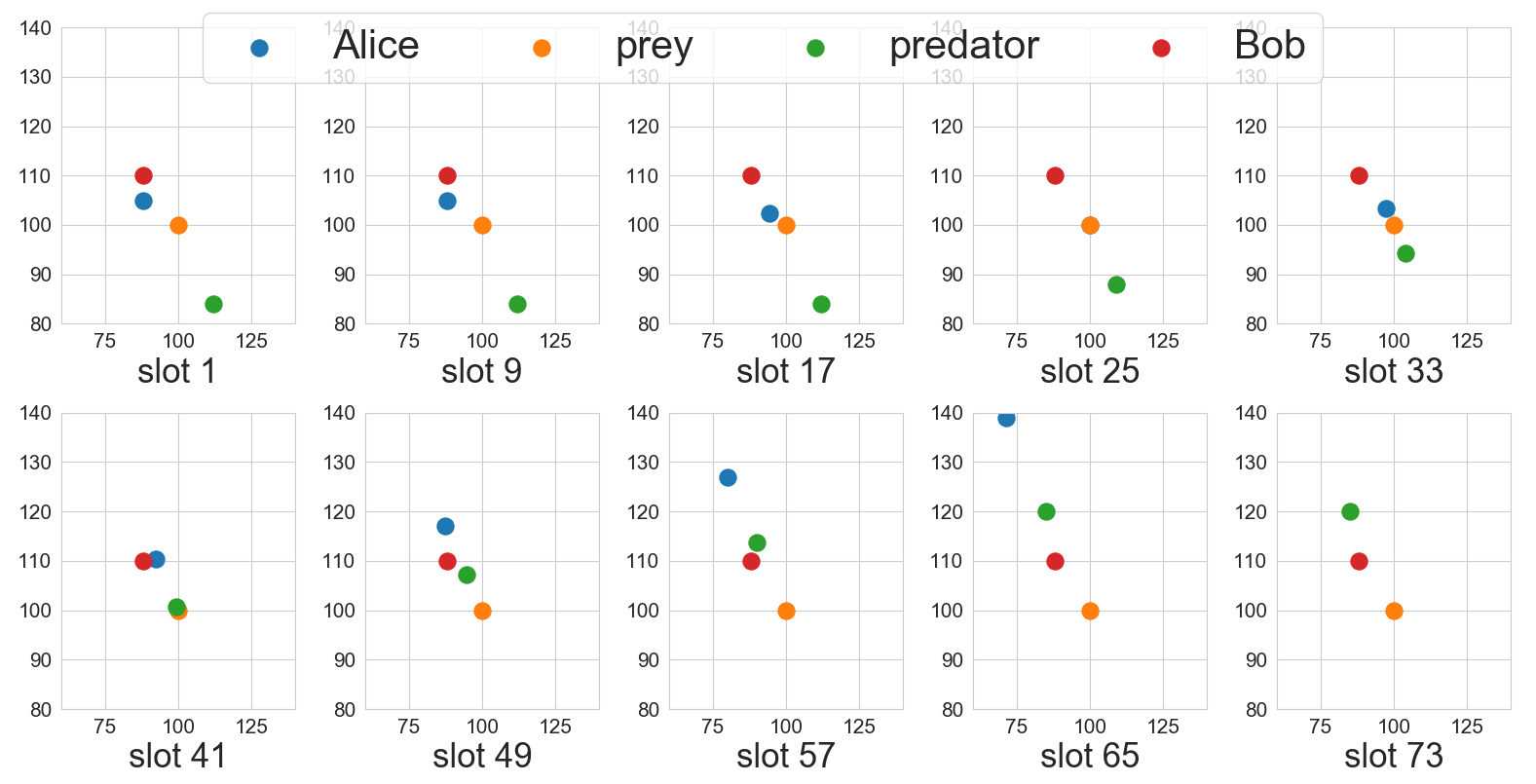}
    \caption{The Status of some Slots in double-agent Experiment}
    \label{fig:exp2full}
\end{figure}

According to the predication based on human, with the remaining of Bob, Alice notices the predator early and get away from the predator early accordingly. When find the predator, Bob reminds Alice at once. Even though Bob is also afraid, he starts keep away from the predator after Alice has a distance from the predator since the friendship need is stronger.

As shown in Figure~\ref{fig:exp2full}, the actual behavior also coincides with the prediction. In the 17th slot, Bob finds the predator and remind Alice. In the 41st slot, the predator gets near to both Alice and Bob. Since at that time, the friend need strength of Bob is higher than safety, he does not choose to flee. In the 49th slot, Alice has flee to a safe place. At that time, the friend need of Bob decreases, and safety becomes the most important need. Bob starts to flee the predator.

The satisfaction is shown in Figure~\ref{fig:exp2}. For Alice, two accidents are to be handled. One is to meet the food need and move to the prey, and the other is flee the predator with Bob's notification. These two accidents corresponds to the 1st and 19th slot. For Bob, more accidents happen. In the 15th slot, his satisfactions of both friendship and safety fluctuate, and then both of them decrease. In the 17th slot, the friendship need goes beyond the threshold, which causes the action that Bob reminds Alice. With the fleeing of Alice, the friendship need satisfaction of Bob increases and gets larger than safety in the 44th slot. This cause the action that when the predator gets too close to Bob during chasing Alice, Bob flees a distance. After Bob gets away from the predator for a distance, he becomes safe. At that time, the friendship need gets stronger than safety, which happens in the 53rd slot.

Note that the ``need of friendship'' for Alice changes even though Bob was close to the predator. The reason is that when Bob is in danger, Alice has been far from him, which means that Bob is out of the scope of the view of Alice.

\section{Conclusions and Future Work}
\label{sec:con}
Conscious AI is an interesting problem that draws attention from the community. In this paper, we develop a preliminary system for conscious AI driven by needs. We implement the STM-LTM framework and a series of LTM algorithms. From the experiments on typical scenarios, our system shows the consciousness as predicted by human, and our system provides explainable mental behavior.

In current version, the prediction model that maps lower level needs to higher level once is trained periodically from the historical data. To avoid the storage of a large amount of the historical data, we will develop an incremental learning model which just store the features that requires for model updating for the historical data. Apart from that, deep think, a specific think method, will be developed for the agent to adapt to complex environments and solve complex problems.


\newpage
\bibliographystyle{named}
\bibliography{ijcai19}

\begin{thebibliography}{}

\bibitem[\protect\citeauthoryear{Baars and Franklin}{2009}]{baars2009}
B.J. Baars and S.~Franklin.
\newblock Consciousness is computational: The lida model of global workspace
  theory.
\newblock {\em International Journal of Machine Consciousness}, 1(1):23--32,
  2009.

\bibitem[\protect\citeauthoryear{Baars}{1988}]{baars1988}
B.J Baars.
\newblock {\em A Cognitive Theory of Consciousness}.
\newblock Cambridge University Press, New York, 1988.

\bibitem[\protect\citeauthoryear{Blum and
  Blum}{2020}]{DBLP:journals/corr/abs-2011-09850}
Manuel Blum and Lenore Blum.
\newblock A theoretical computer science perspective on consciousness.
\newblock {\em CoRR}, abs/2011.09850, 2020.

\bibitem[\protect\citeauthoryear{Blum \bgroup \em et al.\egroup
  }{2019}]{basicmodel}
Manuel Blum, Lenore Blum, and Avrim Blum.
\newblock Towards a conscious ai: A computer architecture inspired by cognitive
  neuroscience.
\newblock Preliminary Draft, 2019.

\bibitem[\protect\citeauthoryear{Blum}{2017}]{blumtalk}
Manuel Blum.
\newblock Can a machine be conscious? towards a computational model of
  consciousness.
\newblock In {\em Academic Talk}, Harbin, China, 2017.

\bibitem[\protect\citeauthoryear{Boogaard \bgroup \em et al.\egroup
  }{2017}]{Boogaard2017}
E~V~D Boogaard, J~Treur, and M~Turpijn.
\newblock A neurologically inspired network model for graziano's attention
  schema theory for consciousness.
\newblock In {\em IWINAC}, 2017.

\bibitem[\protect\citeauthoryear{Chella \bgroup \em et al.\egroup
  }{2018}]{DBLP:conf/aaaiss/2019tocais}
Antonio Chella, David Gamez, Patrick Lincoln, Riccardo Manzotti, and
  Jonathan~D. Pfautz, editors.
\newblock {\em the 2019 Towards Conscious {AI} Systems Symposium co-located
  with the Association for the Advancement of Artificial Intelligence 2019
  Spring Symposium Series {(AAAI} SSS-19), Stanford, CA, March 25-27, 2019},
  volume 2287 of {\em {CEUR} Workshop Proceedings}. CEUR-WS.org, 2018.

\bibitem[\protect\citeauthoryear{Graziano and Webb}{2014}]{Graziano2014}
M~S~A Graziano and T~W Webb.
\newblock A mechanistic theory of consciousness.
\newblock {\em International Journal of Machine Consciousness}, 2014.

\bibitem[\protect\citeauthoryear{Miller}{1956}]{miller1956magical}
George~A Miller.
\newblock The magical number seven, plus or minus two: Some limits on our
  capacity for processing information.
\newblock {\em Psychological review}, 63(2):81, 1956.

\bibitem[\protect\citeauthoryear{Newman and Baars}{1993}]{newman1993}
J.~Newman and B.J. Baars.
\newblock A neural attentional model for access to consciousness: A global
  workspace perspective.
\newblock {\em Concepts in Neuroscience}, 4:255--290, 1993.

\bibitem[\protect\citeauthoryear{Shapiro and Bona}{2010}]{Shapiro2010}
S~C Shapiro and J~P Bona.
\newblock The glair cognitive architecture.
\newblock {\em International Journal of Machine Consciousness}, 2(2):307--332,
  2010.

\bibitem[\protect\citeauthoryear{Taormina and Gao}{2013}]{taormina2013maslow}
Robert~J Taormina and Jennifer~H Gao.
\newblock Maslow and the motivation hierarchy: Measuring satisfaction of the
  needs.
\newblock {\em The American journal of psychology}, 126(2):155--177, 2013.

\bibitem[\protect\citeauthoryear{Thomsen}{2011}]{Thomsen2011}
K~Thomsen.
\newblock Consciousness for the ouroboros model.
\newblock {\em International Journal of Machine Consciousness}, 3(1):239--250,
  2011.

\bibitem[\protect\citeauthoryear{Zhang and
  Wang}{2020}]{DBLP:journals/corr/abs-2003-02446}
Meifan Zhang and Hongzhi Wang.
\newblock {LAQP:} learning-based approximate query processing.
\newblock {\em CoRR}, abs/2003.02446, 2020.

\end{thebibliography}

\end{document}